\documentclass[preprint,5p,times,twocolumn]{elsarticle}
\setcitestyle{square}
\usepackage[T1]{fontenc}
\usepackage[utf8]{inputenc}
\usepackage{microtype}

\usepackage{amsmath}
\usepackage{amssymb}
\usepackage{amsthm}

\usepackage[
  scr=boondox,
  cal=esstix
]{mathalpha}

\usepackage{booktabs}
\usepackage{multirow}
\usepackage{array}
\usepackage{arydshln}

\usepackage{graphicx}
\usepackage{float}
\usepackage{flafter}
\usepackage{subfig}

\usepackage{xcolor}

\definecolor{lightblue}{RGB}{0,176,240}
\definecolor{purple}{RGB}{153,102,255}

\usepackage{enumitem}
\setlist{nosep,leftmargin=*}

\usepackage{algorithm}
\usepackage{algpseudocode}

\usepackage{tikz}
\usetikzlibrary{
  arrows.meta,
  positioning,
  calc,
  fit,
  backgrounds,
  shapes.geometric,
  decorations.pathreplacing
}

\usepackage{pgfplots}
\pgfplotsset{compat=1.18}

\usepackage{framed}
\usepackage{multicol}
\usepackage{nomencl}

\makenomenclature
\renewcommand*\nompreamble{%
  \begin{multicols}{2}%
}

\renewcommand*\nompostamble{%
  \end{multicols}%
}

\usepackage{xspace}
\usepackage{url}

\usepackage[
  colorlinks=true,
  linkcolor=blue!55!black,
  citecolor=green!45!black,
  urlcolor=blue!65!black
]{hyperref}

\hypersetup{
  pdftitle={
    AeroReformer2:
    Spoken-Query Referring Segmentation for Aerial Images
  },
  pdfauthor={Anonymous Authors}
}

\usepackage[
  nameinlink,
  capitalize
]{cleveref}

\crefname{figure}{Figure}{Figures}
\Crefname{figure}{Figure}{Figures}

\crefname{table}{Table}{Tables}
\Crefname{table}{Table}{Tables}

\crefname{section}{Section}{Sections}
\Crefname{section}{Section}{Sections}

\crefname{equation}{Equation}{Equations}
\Crefname{equation}{Equation}{Equations}

\usepackage[switch]{lineno}

\newcommand{\dataset}{VoiceAeroRef\xspace}
\newcommand{\model}{AeroReformer2\xspace}
\newcommand{\svkla}{SV-KLA\xspace}
\newcommand{\cgtm}{CGTM\xspace}

\journal{
  International Journal of Applied Earth Observation
  and Geoinformation
}

\makeatletter
\def\ps@pprintTitle{%
  \let\@oddhead\@empty
  \let\@evenhead\@empty
  \def\@oddfoot{}%
  \let\@evenfoot\@oddfoot
}
\makeatother

\begin{document}

\begin{frontmatter}
\title{AeroReformer2: Spoken-Query Referring Segmentation for Aerial Images}

\author[label1]{Rui Li}
\author[label2]{Chenxi Duan}
\author[label3]{Haoyang Yang\corref{cor2}}
\cortext[cor2]{Corresponding author}

\begin{abstract}
Spoken language offers a natural, hands-free interface for specifying an arbitrary target in dense remote-sensing imagery, yet existing referring remote-sensing image segmentation benchmarks accept only written expressions. To bridge this gap, we introduce \dataset, a spoken-query benchmark derived from RISBench that adds accent- and voice-diverse speech while preserving the original image, mask, and data splits.  Its hard evaluation set combines rotor, wind, and mixed interference with three signal-to-noise levels.  We also propose \model, an efficient bilateral network that combines a boundary-preserving visual path with token-preserving speech encoding, kernel linear cross-modal attention, and a resolution refinement head.  The design conditions visual features at two scales without materializing a dense speech--visual affinity matrix, then restores fine boundaries using high-resolution visual features. On the clean test split, \model with Swin-Base achieves 62.09\% mean intersection over union (mIoU) and 68.22\% overall intersection over union (oIoU), outperforming the strongest audio-adapted remote-sensing baseline by 5.38 and 2.08 percentage points, respectively.  It retains the best hard-set mIoU at 54.09\%.  To the best of our knowledge, this is the first benchmark and model study of full-sentence spoken-query referring segmentation for remote-sensing imagery. 
\end{abstract}

\end{frontmatter}

\section{Introduction}
\label{sec:introduction}

High-resolution aerial and satellite remote sensing supports infrastructure mapping, emergency response, precision agriculture, traffic analysis, environmental monitoring, and large-area Earth observation~\cite{zhang2023efficient, li2025aeroreformer, li2026offshorewake}. Across these applications, pixel-level perception is indispensable: an image-analysis system must not only recognize that roads, buildings, vehicles, vessels, vegetation, or damaged regions exist, but also delineate their precise spatial extent.  Modern semantic segmentation has made this form of scene understanding increasingly reliable~\cite{li2025aeroreformer, long2015fcn,ronneberger2015unet,chen2017deeplab, li2024lswinsr}.  Yet semantic segmentation presumes a fixed class vocabulary and produces every supported category regardless of the analyst's immediate intent.  This is poorly matched to an interactive workflow in which a person may care about one particular object---for example, ``the small vehicle immediately below the bridge,'' ``the vessel at the right-most pier,'' or ``the irregular pool nearest the lower boundary.''  The desired target is defined jointly by its category, appearance, position, and relation to surrounding objects, not by a class label alone.

At the same time, the interface used to express that intent matters.  A point or box prompt is concise, but it requires continuous visual attention and precise cursor placement; its meaning can also remain ambiguous in dense overhead imagery containing many small, adjacent instances.  A fixed command menu avoids ambiguity only by restricting what the user can request.  In contrast, \emph{direct spoken language is a natural interaction modality for an analyst interpreting remote-sensing imagery}: people already describe targets through category words, attributes, landmarks, and spatial relations, and an utterance can express all of these cues in one command while leaving the display available for image inspection.  Speech therefore provides a direct bridge between human intent and pixel-level remote-sensing analysis rather than merely replacing one graphical pointing device with another.

\begin{table*}[t]
\centering
\caption{Positioning of \dataset relative to adjacent datasets. ``Speech'' means that the conditioning signal is a user utterance rather than sound emitted by the target. RRSIS denotes referring remote-sensing image segmentation, AVS denotes audio-visual segmentation, AVOS denotes audio-guided video object segmentation, and IS denotes image segmentation. Counts use annotated frames/targets or triplets as reported by the corresponding work and are therefore not directly comparable across tasks.}
\label{tab:dataset_comparison}
\footnotesize
\setlength{\tabcolsep}{3.1pt}
\begin{tabular}{lllrrll}
\toprule
Dataset & Domain & Unit & Scale & Pixel mask & Speech & Query form \\
\midrule
RefSegRS~\cite{yuan2024rrsis} & Remote sensing & image & 4,420 & Yes & No & full expression \\
RRSIS-D~\cite{liu2024rmsin} & Remote sensing & image & 17,402 & Yes & No & full expression \\
RISBench~\cite{dong2024crobim} & Remote sensing & image & 52,472 & Yes & No & full expression \\
NWPU-Refer~\cite{yang2025nwpurefer} & Remote sensing & image & 49,745 & Yes & No & full expression \\
AVSBench~\cite{zhou2022avs} & Natural scene & video & -- & Yes & No & target sound \\
Ref-AVS~\cite{wang2024refavs} & Natural scene & video & -- & Yes & No & multimodal cues \\
Wnet/AVOS~\cite{pan2022wnet} & Natural scene & video & -- & Yes & Yes & full expression \\
Audio-guided IS~\cite{santos2025audioguided} & Natural scene & image & 66,202 & Yes & Yes & single word \\
OmniAVS~\cite{ying2025omniavs} & Natural scene & video & 2,104 & Yes & Yes & multimodal prompt \\
\textbf{\dataset (ours)} & \textbf{Remote sensing} & \textbf{image} & \textbf{52,466} & Yes & \textbf{Yes} & \textbf{full expression} \\
\bottomrule
\end{tabular}
\end{table*}

Referring image segmentation (RIS) provides the dense-prediction formulation needed for that bridge.  Given an image and a free-form expression, RIS predicts exactly the pixels belonging to the described referent~\cite{hu2016segmentation,liu2017recurrent,li2018rrn}.  Unlike semantic segmentation, the task can distinguish multiple objects of the same class and accommodate previously unenumerated combinations of attributes and relations.  Recent cross-modal attention, vision-language pretraining, and query-based methods have substantially improved grounding in natural images~\cite{ye2019cmsa,huang2020cmpc,yang2022lavt,wang2022cris,hu2023dmmi,shah2024lqmformer}.  Referring remote-sensing image segmentation (RRSIS) transfers this paradigm to overhead imagery, where targets can be only a few pixels wide, appear at arbitrary orientations, span extreme scales, and require long-range geospatial context~\cite{li2025aeroreformer, yuan2024rrsis,liu2024rmsin,dong2024crobim}.  RISBench is a particularly valuable resource: it contains 52,472 image--expression--mask triplets assembled from the Dataset for Object Detection in Aerial Images version 2 (DOTA-v2) and Dataset for Object deTection in Optical Remote sensing images (DIOR)~\cite{dong2024crobim,xia2018dota,li2020dior}.

However, existing RRSIS benchmarks provide written text rather than a spoken instruction.  Treating a transcript as equivalent to speech removes precisely the variability encountered by a deployable voice interface.  The same sentence changes acoustically with accent, gender presentation, pitch, rate, microphone response, and duration.  Recognition-relevant evidence is distributed across a sequence of waveform frames rather than delivered as symbolic tokens.  In practical image-analysis environments, the speech channel may additionally be affected by machinery, ventilation, wind, microphone response, and changing signal-to-noise ratio.  A text-only benchmark cannot measure sensitivity to these acoustic variables or determine whether they introduce failures beyond the visual and linguistic errors already present in text-based grounding.  Full-sentence speech does not itself solve relational reasoning; rather, it tests whether cues such as ``left-most,'' ``between,'' and ``next to'' remain usable after acoustic encoding.

Prior speech-conditioned segmentation establishes the feasibility of grounding verbal instructions, but it does not address aerial images.  Wnet introduced audio-guided video object segmentation from spoken referring expressions~\cite{pan2022wnet}, and STBridge studied noise-tolerant speech-referring video object segmentation~\cite{li2023speechrvos}.  More recently, direct audio-guided image segmentation has been evaluated with single-word spoken queries~\cite{santos2025audioguided}, while OmniAVS combines speech, sound, text, and visual cues in videos~\cite{ying2025omniavs}.  These studies concern natural-image or video scenes rather than full-sentence references to small, repeated, and arbitrarily oriented aerial targets.  The remaining gap is a benchmark that jointly provides aerial imagery, complete spoken referring expressions, pixel masks, and controlled speaker diversity.

Operational use also makes efficiency a first-class research constraint.  Dense cross-modal attention over $N$ visual positions and $T$ speech tokens materializes an $N\times T$ affinity matrix.  For high-resolution remote-sensing inputs and multi-second utterances, that matrix competes with the visual backbone for limited accelerator memory, while quadratic visual self-attention can further increase cost.  Efficient bilateral segmenters preserve a high-resolution detail stream while computing semantic context at lower resolution~\cite{yu2018bisenet,li2021abcnet,fan2021stdc}.  Linear-attention formulations rearrange attention products so that the intermediate state depends on channel dimension rather than the product of token counts~\cite{shen2021efficient,katharopoulos2020linear}.  This combination is especially attractive for spoken-query remote-sensing segmentation: the boundary path protects tiny objects, while token-preserving linear fusion can ground a variable-length utterance without storing a dense speech--visual affinity map.

Motivated by these observations, we introduce \dataset and \model.  \dataset converts the expressions of RISBench into a controlled speech benchmark without altering image or mask splits.  Training expressions receive eight accent--gender variants; at every epoch, each source sample draws one of these recordings so that voice diversity does not create an artificial eightfold epoch.  Validation and test use category-balanced single voices.  \model couples a bilateral visual encoder with wav2vec~2.0 speech tokens~\cite{baevski2020wav2vec}, a learned token memory, and dual-scale kernel linear cross-attention.  Rather than presenting bilateral segmentation, token gating, or linear attention as isolated inventions, the architecture integrates them around the particular geometry of spoken-query remote-sensing segmentation: associative speech--visual fusion preserves the token sequence, reliability weights enter the sufficient statistics used by every spatial query, coarse and fine fusion resolve localization at two scales, and a parallel detail path retains small-object boundaries.

\begin{figure*}[t]
\centering
  \includegraphics[width=\linewidth]{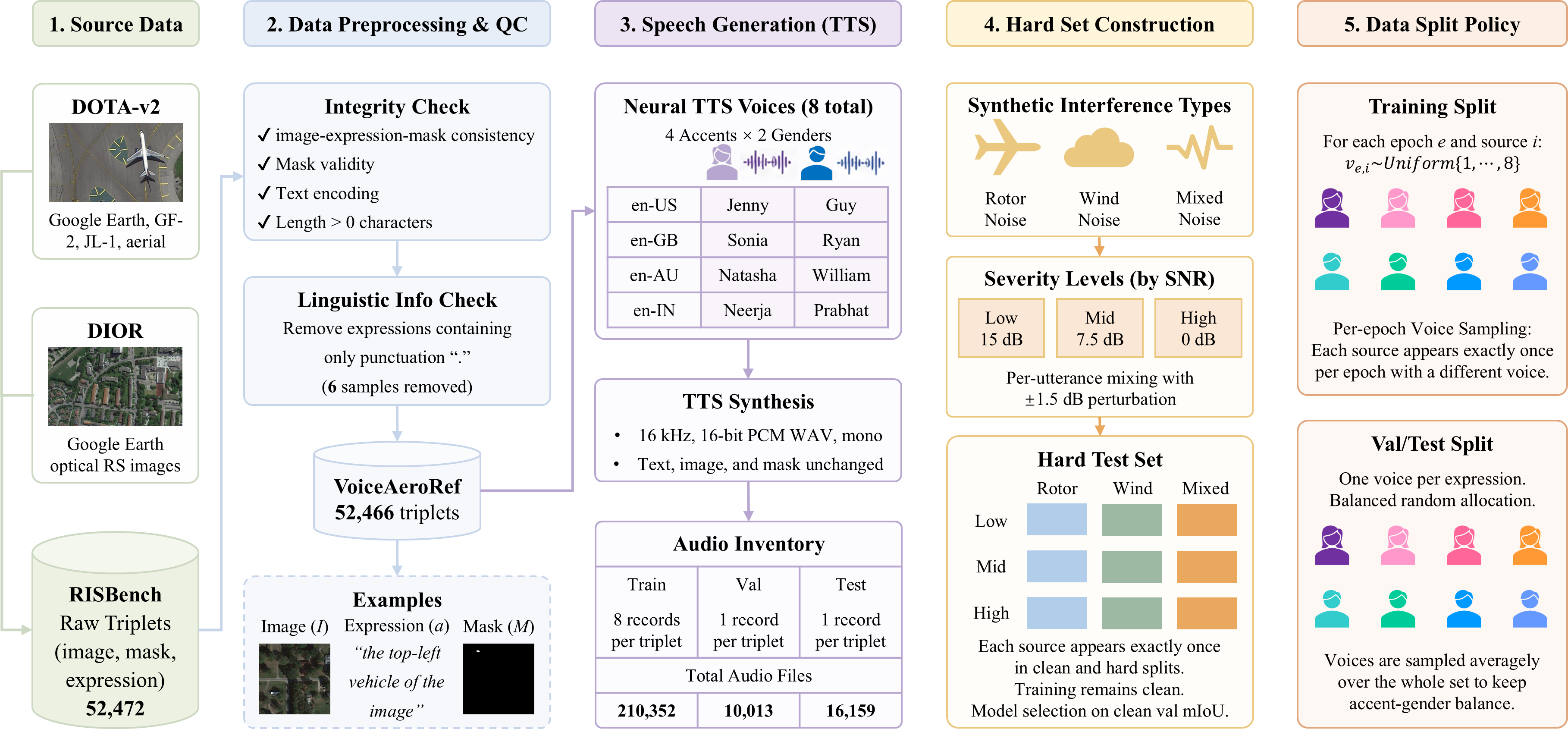}
\caption{\dataset dataset construction and evaluation protocol. The pipeline covers source-data quality control, controlled neural speech synthesis, hard-test construction, and split-specific voice sampling.}
\label{fig:dataset_pipeline}
\end{figure*}

To the best of our knowledge, this is the first study to formulate and benchmark full-sentence spoken-query referring segmentation for remote-sensing imagery.  This work makes four contributions:
\begin{itemize}
  \item We formulate full-sentence spoken-query referring segmentation for remote-sensing imagery and construct \dataset from RISBench through systematic, split-preserving quality control.
  \item We establish a controlled speech protocol with eight accent--gender recordings for every training expression, balanced single-voice clean evaluation, and a nine-condition hard evaluation grid; one candidate is sampled per source sample per epoch.
  \item We propose \model, an efficient speech-conditioned bilateral network with token-preserving speech encoding, confidence-gated token memory (CGTM), and dual-scale speech-visual kernel linear attention (SV-KLA) that avoids a dense visual-token--speech-token affinity matrix.
  \item On the 16,159-sample test split, the Swin-Base version reaches 62.09\% clean mIoU and 54.09\% hard mIoU, exceeding the strongest non-ours method by 5.38 and 5.63 percentage points, respectively.
\end{itemize}

\section{Related Work}
\label{sec:related}

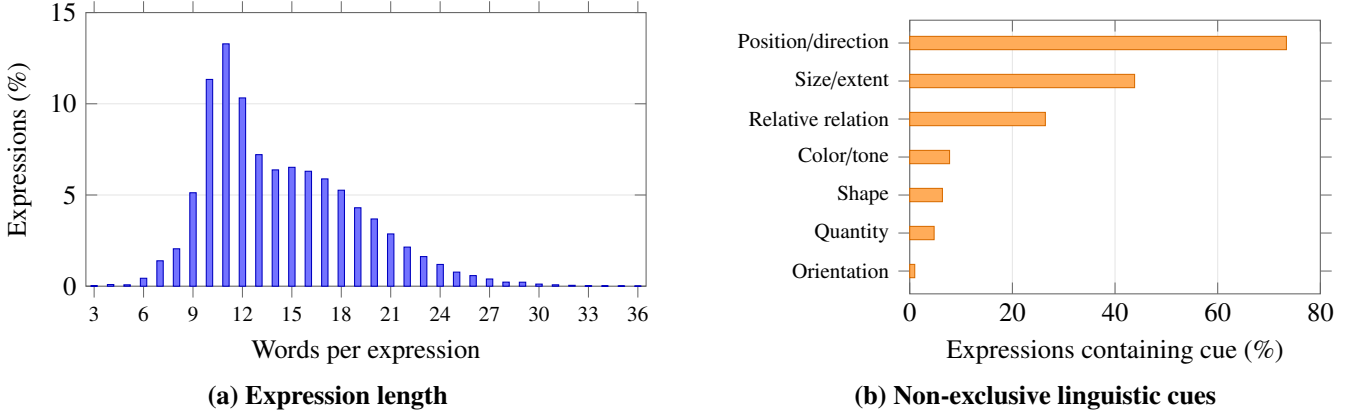
\begin{figure*}[t]
\centering
\begin{minipage}[t]{0.49\linewidth}
\centering
\begin{tikzpicture}
\begin{axis}[
  width=\linewidth,height=52mm,
  ybar,bar width=2.4pt,
  xmin=2.5,xmax=36.5,
  ymin=0,ymax=15,
  ylabel={Expressions (\%)},
  xlabel={Words per expression},
  xtick={3,6,9,12,15,18,21,24,27,30,33,36},
  x tick label style={font=\footnotesize},
  ymajorgrids=true,grid style={black!10},
  axis line style={black!60},tick style={black!60}
]
\addplot[fill=blue!55,draw=blue!75!black] coordinates {
  (3,0.025) (4,0.091) (5,0.076) (6,0.435) (7,1.393) (8,2.051)
  (9,5.123) (10,11.335) (11,13.285) (12,10.321) (13,7.212)
  (14,6.376) (15,6.517) (16,6.301) (17,5.882) (18,5.262)
  (19,4.300) (20,3.686) (21,2.865) (22,2.144) (23,1.624)
  (24,1.193) (25,0.772) (26,0.581) (27,0.395) (28,0.223)
  (29,0.219) (30,0.116) (31,0.076) (32,0.048) (33,0.023)
  (34,0.017) (35,0.015) (36,0.008)
};
\end{axis}
\end{tikzpicture}

\textbf{(a) Expression length}
\end{minipage}\hfill
\begin{minipage}[t]{0.49\linewidth}
\centering
\begin{tikzpicture}
\begin{axis}[
  width=0.78\linewidth,height=52mm,
  xbar,bar width=5pt,
  xmin=0,xmax=80,
  xlabel={Expressions containing cue (\%)},
  ytick={1,2,3,4,5,6,7},
  yticklabels={Orientation,Quantity,Shape,Color/tone,Relative relation,Size/extent,Position/direction},
  y tick label style={font=\footnotesize},
  xmajorgrids=true,grid style={black!10},
  axis line style={black!60},tick style={black!60}
]
\addplot[fill=orange!65,draw=orange!85!black] coordinates {
  (0.976,1) (4.752,2) (6.372,3) (7.761,4) (26.425,5) (43.809,6) (73.390,7)
};
\end{axis}
\end{tikzpicture}

\textbf{(b) Non-exclusive linguistic cues}
\end{minipage}
\caption{Linguistic characteristics of the usable expressions. Panel (a) gives the frequency of each expression length from 3 to 36 words; only five expressions are longer than 36 words. Panel (b) reports the prevalence of lexicon-based linguistic cues. Cue categories are non-exclusive and therefore do not sum to 100\%.}
\label{fig:language_stats}
\end{figure*}

\subsection{Referring image and remote-sensing segmentation}

RIS was introduced as pixel-level grounding of a natural-language expression.  Early systems concatenated convolutional image features, spatial coordinates, and recurrent sentence representations~\cite{hu2016segmentation}; recurrent refinement then enabled iterative multimodal reasoning~\cite{liu2017recurrent}, while the recurrent refinement network (RRN) modeled cross-modal relationships at multiple visual levels~\cite{li2018rrn}.  Attention-based methods replaced a single global sentence vector with finer word--region interaction.  The cross-modal self-attention network (CMSA) constructed self-attention over multimodal word--pixel features~\cite{ye2019cmsa}; cross-modal progressive comprehension (CMPC) progressively understood entity and attribute cues~\cite{huang2020cmpc}; and linguistic-structure context modeling used phrase structure to guide visual reasoning~\cite{hui2020lscm}.  These works established two recurring requirements: the decoder needs local detail for accurate masks, and the fusion mechanism must preserve enough language structure to distinguish similar instances.

Transformers and large-scale vision-language pretraining moved fusion earlier and made it more expressive.  The language-aware vision transformer (LAVT) injects linguistic information throughout a hierarchical visual transformer~\cite{yang2022lavt}, whereas CLIP-driven referring image segmentation (CRIS) transfers contrastive language--image pretraining (CLIP) knowledge through a vision-language decoder and text-to-pixel contrastive learning~\cite{wang2022cris}.  Dynamic multimodal interaction (DMMI) reconsiders one-expression-to-one-object assumptions and performs dynamic multimodal reasoning~\cite{hu2023dmmi}.  The language-aware query mask transformer (LQMFormer) addresses query collapse through language-aware feature fusion and dynamic query selection~\cite{shah2024lqmformer}.  Other recent directions include global--local zero-shot grounding~\cite{yu2023zeroshot}, prompt-driven adaptation of CLIP and the Segment Anything Model (SAM)~\cite{kirillov2023sam,shang2024promptris}, explicit mask grounding supervision~\cite{chng2024magnet}, feature interfaces that bridge SAM and CLIP~\cite{ito2025feature}, and training-free hybrid global--local mask representations~\cite{liu2025hybrid}.  These methods improve accuracy, supervision efficiency, or open-vocabulary transfer, but they assume written text and generally do not optimize the speech-token--pixel interaction for edge deployment.

RRSIS adds domain-specific difficulties that natural-image RIS benchmarks underrepresent.  RefSegRS introduced 4,420 remote-sensing image--expression--mask triplets together with language-guided cross-scale enhancement~\cite{yuan2024rrsis}.  RRSIS-D and the rotated multi-scale interaction network (RMSIN) emphasize rotated targets and multi-scale interaction~\cite{liu2024rmsin}.  RISBench scales the task to 52,472 triplets, and the cross-modal bidirectional interaction model (CroBIM) improves cross-modal bidirectional interaction~\cite{dong2024crobim}; the source expressions originate from the broader Versatile Vision-Language Benchmark for Remote Sensing Image Understanding (VRSBench) collection~\cite{li2024vrsbench}.  AeroReformer addresses referring segmentation in a separate low-altitude aerial setting through vision--language cross-attention and rotation-aware multi-scale fusion~\cite{li2025aeroreformer}.  Recent models refine this interaction in complementary ways: FIANet performs fine-grained image--text alignment~\cite{lei2025fianet}; the scale-wise bidirectional alignment network (SBANet) adds bidirectional alignment across feature scales~\cite{li2025sbanet}; STDNet couples spatial multi-scale correlation with target/background decoding~\cite{zhang2025stdnet}; and long-term semantic-guidance ConvFormer (LSCF) maintains language guidance through its decoder~\cite{ma2025lscf}.  RSSep investigates joint referring segmentation and caption generation~\cite{ho2024rssep}, while NWPU-Refer expands geographical diversity and high-resolution coverage~\cite{yang2025nwpurefer}.  Despite rapid progress, all of these resources deliver text.  None exposes accent, speaker, or duration as benchmark variables, so none can separate acoustic encoding from visual grounding quality.

\subsection{Efficient segmentation and linear attention}

Encoder--decoder networks such as fully convolutional networks (FCNs), U-Net, and DeepLab established strong dense-prediction baselines~\cite{long2015fcn,ronneberger2015unet,chen2017deeplab}, but high-resolution remote-sensing inference requires more deliberate accuracy--latency trade-offs.  MobileNetV3 with Lite Reduced Atrous Spatial Pyramid Pooling (LR-ASPP) uses hardware-aware mobile blocks~\cite{howard2019mobilenetv3}.  The bilateral segmentation network (BiSeNet) separates spatial detail from semantic context~\cite{yu2018bisenet}; the Short-Term Dense Concatenate network (STDC) revisits the cost of a separate spatial path and adds detail supervision~\cite{fan2021stdc}; the proportional--integral--derivative network (PIDNet) uses three corresponding branches~\cite{xu2023pidnet}.  For fine-resolution remote sensing, the Attentive Bilateral Contextual Network (ABCNet) combines an attentive context path with a shallow spatial branch~\cite{li2021abcnet}.  \model retains this bilateral principle because tiny overhead objects need early high-resolution evidence, but replaces late text conditioning with token-level speech fusion.

Standard dot-product attention forms pairwise affinities before normalizing values~\cite{vaswani2017attention}.  A visual self-attention layer stores $O(N^2)$ affinities, and speech--visual cross-attention stores $O(NT)$ affinities for $N$ pixels and $T$ speech tokens.  Efficient attention shows that separating and reordering key--value products can yield linear complexity under suitable normalization~\cite{shen2021efficient}; kernel linear transformers use a positive feature map and compute $\phi(Q)(\phi(K)^\top V)$ associatively~\cite{katharopoulos2020linear}.  In high-resolution aerial image super-resolution, the Linear Swin Transformer for Super-Resolution (LSwinSR) further demonstrates that kernelized attention can reduce the computational burden of Swin-style restoration while retaining competitive reconstruction accuracy~\cite{li2024lswinsr}.  This distinction is architectural rather than merely an optimized implementation: the dense affinity tensor is never instantiated.  \model specializes the formulation to dense speech grounding at two spatial scales and learns a confidence weight for each speech token because the tokens contribute unequally to identifying the referent.

\begin{table*}[t]
\centering
\caption{Dataset splits and audio counts after quality control.  Hard audio is derived only for the test split; images, expressions, masks, and clean voice assignments are unchanged.}
\label{tab:split_stats}
\small
\begin{tabular}{lrrrr}
\toprule
Split & Usable sources & Clean voices/source & Clean audio & Hard audio \\
\midrule
Train & 26,294 & 8 & 210,352 & -- \\
Validation & 10,013 & 1 & 10,013 & -- \\
Test & 16,159 & 1 & 16,159 & 16,159 \\
\midrule
Total & 52,466 & -- & 236,524 & 16,159 \\
\bottomrule
\end{tabular}
\end{table*}

\subsection{Audio-visual segmentation and speech representation}

Audio-visual segmentation (AVS) conventionally localizes pixels associated with sounding objects.  AVSBench introduced single-source, multi-source, and semantic settings with temporal pixel-wise interaction~\cite{zhou2022avs}.  AVSegFormer uses dense audio-visual mixing and transformer decoding~\cite{gao2024avsegformer}; annotation-free AVS composes image-mask and audio data and adapts SAM efficiently~\cite{liu2024annotationfree}; Cooperation of Multi-order Bilateral Relations (COMBO) models bilateral relations at pixel, modality, and temporal levels~\cite{yang2024combo}.  Ref-AVS uses multimodal scene cues to select objects in videos~\cite{wang2024refavs}, and OmniAVS extends this formulation to expressions that can combine text, speech, sound, and visual prompts~\cite{ying2025omniavs}.

Speech-referring segmentation is more directly related to our task.  Wnet introduced an audio-guided video segmentation benchmark and an end-to-end model for spoken referring expressions~\cite{pan2022wnet}.  STBridge aligns speech and text representations to improve noise tolerance in referring video object segmentation~\cite{li2023speechrvos}.  Santos \emph{et al.} subsequently benchmarked direct audio-guided segmentation of static natural images, but restricted the spoken instruction to a single object keyword~\cite{santos2025audioguided}.  These works show that speech can serve as the semantic query rather than as environmental evidence.  They do not, however, study full relational sentences, aerial imagery, or the scale and orientation variation of remote-sensing targets.  \dataset addresses this intersection while retaining the original RISBench masks and full referring expressions.

Self-supervised speech models provide a more appropriate representation for this setting than a global environmental-audio embedding.  wav2vec~2.0 learns contextualized representations directly from masked waveform latents~\cite{baevski2020wav2vec}; HuBERT uses clustered hidden-unit prediction~\cite{hsu2021hubert}; and WavLM extends self-supervision toward full-stack speech processing under noisy and overlapping conditions~\cite{chen2022wavlm}.  SpecAugment demonstrates the broader value of time--frequency masking for robust speech recognition~\cite{park2019specaugment}.  We adopt wav2vec~2.0 because its frame sequence can be retained for word-level visual grounding.  Attributes and relations may occupy brief intervals and differ in task relevance, so the sequence is not collapsed to a single sentence vector before fusion.

\Cref{tab:dataset_comparison} situates \dataset among adjacent aerial and audio-guided benchmarks.  Existing remote-sensing referring datasets provide complete written expressions but no speech, whereas prior speech-conditioned datasets focus on natural-scene images or videos and often use single words or multimodal prompts.  \dataset is distinguished by the conjunction of the remote-sensing domain, pixel masks, full spoken expressions, and eight controlled accent--gender categories at the scale of RISBench.

\section{\dataset Dataset}
\label{sec:dataset}

\subsection{Task definition}

Each sample is a triplet $(I,a,M)$, where $I\in\mathbb{R}^{H\times W\times3}$ is an aerial or remote-sensing image, $a\in\mathbb{R}^{L}$ is a spoken referring expression sampled at 16 kHz, and $M\in\{0,1\}^{H\times W}$ is the binary mask of the referred object.  The model predicts
\begin{equation}
  \widehat M = f_{\theta}(I,a),
\end{equation}
and is evaluated on its overlap with $M$.  Unlike sound-source segmentation, the utterance need not be acoustically related to the object; its linguistic content provides the reference.

\subsection{Source data and quality control}

RISBench is derived from DOTA-v2 and DIOR imagery and contains 512$\times$512 images, referring expressions, and human-verified masks~\cite{dong2024crobim}. DOTA-v2 combines Google Earth imagery, Gaofen-2 (GF-2) and Jilin-1 (JL-1) satellite imagery, and other aerial imagery, whereas DIOR is a benchmark of optical remote-sensing images collected from Google Earth~\cite{xia2018dota,li2020dior}.  RISBench is therefore a mixed-source overhead remote-sensing dataset. Its released splits contain 26,300 training, 10,013 validation, and 16,159 test records, totaling 52,472 triplets.

We verify record integrity and linguistic informativeness before speech synthesis.  Six training expressions contain only the punctuation mark ``.'' and therefore provide no referring content; excluding them yields 52,466 usable triplets.  As shown in \cref{fig:dataset_pipeline}, the construction preserves every valid image--expression--mask association while adding controlled speech realizations.  The training split stores eight recordings per expression but retains 26,294 effective samples per epoch, whereas validation and test contain one clean recording per expression.  The hard test set reuses the clean test voice assignments and adds one controlled interference condition to every utterance. The resulting split sizes and audio inventory are summarized in \cref{tab:split_stats}.

\begin{figure*}[t]
\centering
\includegraphics[width=\linewidth]{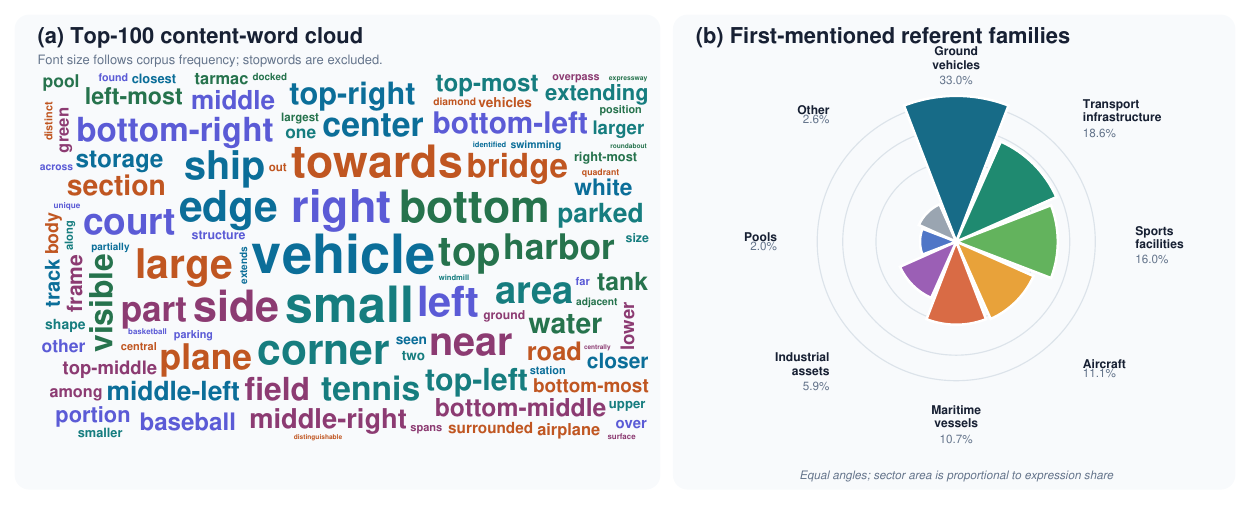}
\caption{Lexical and referent-semantic composition of the 52,466 usable expressions. Panel (a) shows the 100 most frequent non-stopword tokens, with font size proportional to log frequency. Panel (b) is a Nightingale rose of first-mentioned referent families; sector area is proportional to expression share. ``Other'' combines buildings, natural/water terms, and unmatched expressions.}
\label{fig:lexical_semantic}
\end{figure*}

\subsection{Speech generation and split policy}

We synthesize English speech using eight neural text-to-speech (TTS) voices arranged as a $4\times2$ accent--gender factorial design (\cref{tab:voices}). Each output is stored as a mono, 16-bit PCM WAV file sampled at 16 kHz. TTS synthesis changes the acoustic realization but never changes the text, image, mask, or split.

For training, every usable description has eight candidate recordings.  The dataset is indexed by the 26,294 source descriptions, not by the 210,352 audio files.  At epoch $e$, source index $i$ selects
\begin{equation}
  v_{e,i} \sim \operatorname{Uniform}\{1,\ldots,8\},
  \label{eq:voice_sampling}
\end{equation}
with the voice resampled at each epoch.  Thus every source occurs exactly once in an epoch while its acoustic realization varies over training.  This provides speaker augmentation without oversampling identical visual masks by a factor of eight.

Validation and test use one voice per description.  The eight voice categories are assigned by balanced random allocation, so the count of any two categories differs by at most one.  Validation contains 1,251--1,252 utterances per category and test contains 2,019--2,020; accent and gender marginals are therefore balanced.

\begin{table}[t]
\centering
\caption{Controlled neural text-to-speech (TTS) voices used to synthesize the spoken queries.}
\label{tab:voices}
\small
\begin{tabular}{llll}
\toprule
Accent & Locale & Female voice & Male voice \\
\midrule
United States & en-US & JennyNeural & GuyNeural \\
Great Britain & en-GB & SoniaNeural & RyanNeural \\
Australia & en-AU & NatashaNeural & WilliamNeural \\
India & en-IN & NeerjaNeural & PrabhatNeural \\
\bottomrule
\end{tabular}
\end{table}

\subsection{Hard evaluation construction}
\label{sec:hard_construction}

In plausible field or mobile operating scenarios, an analyst or operator may issue spoken queries outdoors or near active equipment, where the microphone is exposed to wind, rotor/machinery noise, and rapidly changing signal-to-noise ratios. We therefore construct a hard evaluation set to test whether spoken referring expressions remain usable under representative acoustic interference, without changing the underlying visual or linguistic task.

The hard test set evaluates acoustic interference without changing the visual or linguistic task.  Each clean evaluation utterance is mixed with locally synthesized rotor noise, wind noise, or a 0.6/0.4 mixture of the two.  Rotor interference combines a four-source harmonic model, amplitude beating, and colored mechanical noise; wind interference combines low-pass colored noise, a time-varying gust envelope, and low-frequency rumble.  The construction uses no external noise recordings and does not model the acquisition platform of the source imagery.

We cross the three interference types with low, medium, and high severity, defined by nominal signal-to-noise ratios of 15, 7.5, and 0 decibels (dB).  A per-utterance perturbation sampled within $\pm1.5$ dB prevents each severity from collapsing to one fixed mixing ratio.  The nine conditions are allocated as evenly as possible: each test cell contains 1,795--1,796.  Every source appears exactly once in the hard test split, which therefore retains 16,159 samples rather than expanding evaluation ninefold. Training remains clean and the reported hard test results use checkpoints selected on clean validation mIoU.

\subsection{Linguistic statistics}
\label{sec:ling_stats}

The 52,466 usable expressions contain 751,126 lowercase regex-tokenized word occurrences.  Mean length is 14.316 words (median 13, standard deviation 4.505, 90th percentile 21, range 3--46).  The normalized vocabulary contains 3,055 types.  The difference from the 4,431-word vocabulary reported for RISBench~\cite{dong2024crobim} arises from lowercase regex tokenization and normalization.

As shown in \cref{fig:language_stats}, expression lengths concentrate around 10--18 words, while a substantial tail extends beyond 20 words.  Position/direction cues appear in 73.390\% of expressions and size/extent cues in 43.809\%, confirming that aerial referring expressions depend strongly on relational and scale language.  Relative relations occur in 26.425\%; color/tone, shape, quantity, and orientation occur in 7.761\%, 6.372\%, 4.752\%, and 0.976\%, respectively.  These categories are detected with transparent keyword lexicons and may overlap.  The most frequent normalized content words include \emph{vehicle} (17,590), \emph{small} (12,867), \emph{right} (8,900), \emph{towards} (8,520), \emph{bottom} (8,502), \emph{side} (7,606), \emph{edge} (7,321), and \emph{corner} (7,298).  This distribution motivates token-level grounding: pooling the utterance into a single vector can blur the object noun, attribute, and spatial relation that jointly identify the target.

To characterize referent semantics, each expression is assigned to the family of its earliest matching referent keyword because RISBench does not provide explicit object-category labels.  As shown in \cref{fig:lexical_semantic}, the word cloud is dominated by object nouns such as \emph{vehicle}, \emph{ship}, \emph{court}, and \emph{plane}, together with spatial terms such as \emph{right}, \emph{bottom}, and \emph{corner}.  Ground vehicles form the largest first-mentioned referent family (33.0\%), followed by transport infrastructure (18.6\%) and sports facilities (16.0\%).  Aircraft and maritime vessels contribute a further 11.1\% and 10.7\%, respectively, whereas industrial assets (5.9\%), pools (2.0\%), and the combined other group (2.6\%) are less frequent.  This long-tailed semantic composition complements the cue analysis: the model must retain fine-grained noun evidence for minority targets while using position and relation words to separate repeated instances in the dominant vehicle and infrastructure scenes.

\section{AeroReformer2}
\label{sec:method}

\begin{figure*}[t]
\centering
\includegraphics[width=\linewidth]{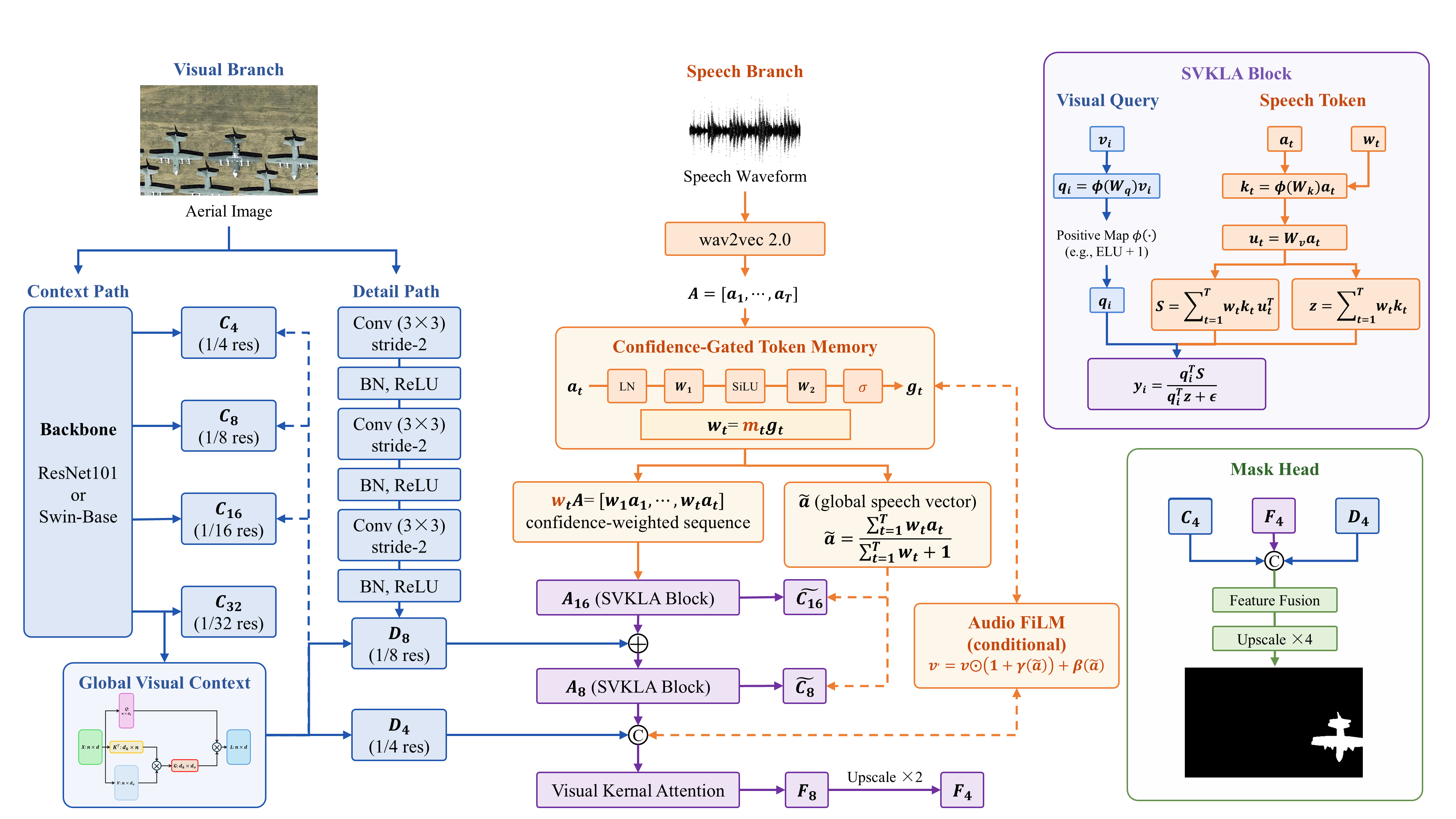}
\caption{\model architecture for spoken-query aerial referring segmentation. The model combines bilateral visual encoding, confidence-gated speech representations, dual-scale speech–visual kernel linear attention, and high-resolution mask refinement.}
\label{fig:labcnet}
\end{figure*}

\model denotes \emph{Aerial Referring Transformer 2}.  The model inherits the bilateral efficiency principle of ABCNet~\cite{li2021abcnet}: a shallow detail path retains boundary evidence at $1/8$ resolution, while a hierarchical context path exposes semantic features at $1/4$, $1/8$, $1/16$, and $1/32$.  We instantiate the context path with either ResNet-101 or Swin-Base; the speech fusion and decoder are identical across both versions.  Depthwise-separable convolutions are used in the detail paths, top-down refinement, fusion blocks, and mask heads.  Kernel linear self-attention at $1/32$ provides global visual context without an $N\times N$ spatial affinity matrix.

The speech path transforms a waveform into a sequence $A=[a_1,\ldots,a_T]\in\mathbb{R}^{T\times D_a}$ using wav2vec~2.0~\cite{baevski2020wav2vec}.  Unlike sentence-level feature-wise linear modulation (FiLM) baselines, \model does not discard the token sequence before fusion.  \cgtm assigns a learned confidence to each token, and the same confidence-weighted sequence conditions context features through \svkla at $1/16$ for coarse semantic localization and $1/8$ for fine grounding.  The conditioned context is fused with the $1/8$ detail path and globally refined by visual kernel attention.  A subsequent $1/4$ head combines this fused semantic feature with top-down backbone context and a separate high-resolution detail path before producing the binary mask.  \Cref{fig:labcnet} summarizes the dataflow.

The resulting architecture is tailored to spoken referring segmentation rather than a simple substitution of one attention operator for another.  Associative kernel attention keeps the acoustic token sequence available to every visual query without constructing a dense token--pixel affinity tensor; the learned token weights enter both the kernel sufficient statistics and the global conditioning branch; and the two speech-fusion scales separate coarse referent selection from fine grounding, after which the $1/4$ visual refinement stage restores boundaries without expanding cross-modal attention to that grid.  These choices form one connected routing strategy for the repeated objects, long spatial relations, and fine object boundaries common in aerial imagery.

\begin{table*}[t]
\centering
\caption{Test-set accuracy on \dataset (\%). The metrics are mean intersection over union (mIoU), overall intersection over union (oIoU), and precision at the indicated intersection-over-union threshold (Pr@.). All methods use the same spoken-query input and 0.5 mask threshold. The best and second-best values are shown in bold and underlined, respectively. ``Adapter'' denotes an audio-token adaptation of the cited text-guided RRSIS model; CNN denotes convolutional neural network.}
\label{tab:main_accuracy}
\footnotesize
\setlength{\tabcolsep}{3.1pt}
\begin{tabular}{llrrrrrrr}
\toprule
Method & Backbone & mIoU & oIoU & Pr@.5 & Pr@.6 & Pr@.7 & Pr@.8 & Pr@.9 \\
\midrule
Audio U-Net & CNN & 27.84 & 36.10 & 21.12 & 15.03 & 10.32 & 6.26 & 2.28 \\
Audio FCN & ResNet-101 & 49.54 & 60.17 & 55.13 & 48.64 & 40.74 & 29.45 & 13.65 \\
Audio DeepLabV3 & ResNet-101 & 51.87 & 62.17 & 57.89 & 51.49 & 43.55 & 32.40 & 15.90 \\
\midrule
FIANet adapter~\cite{lei2025fianet} & Swin-B & 53.48 & 63.86 & 58.99 & 53.15 & 46.59 & 36.52 & 20.43 \\
SBANet adapter~\cite{li2025sbanet} & Swin-B & 55.28 & 65.03 & 61.63 & 55.37 & 48.04 & 37.48 & 20.76 \\
STDNet adapter~\cite{zhang2025stdnet} & Swin-B & 50.34 & 61.22 & 53.36 & 46.96 & 40.17 & 31.36 & 17.40 \\
LSCF adapter~\cite{ma2025lscf} & Swin-B & 56.71 & 66.14 & 63.33 & 57.28 & 49.68 & 39.67 & 22.19 \\
\midrule
\model-ResNet101 (ours) & ResNet-101 & \underline{59.26} & \underline{66.23} & \underline{66.36} & \underline{60.76} & \underline{53.85} & \underline{43.34} & \underline{24.49} \\
\textbf{\model-Swin-B (ours)} & Swin-B & \textbf{62.09} & \textbf{68.22} & \textbf{69.25} & \textbf{63.98} & \textbf{56.71} & \textbf{46.15} & \textbf{27.49} \\
\bottomrule
\end{tabular}
\end{table*}

\subsection{Confidence-Gated Token Memory}

Speech tokens do not contribute equally to referring segmentation.  Object nouns, attributes, and spatial relations often carry more task-relevant evidence than function words or redundant frames.  \cgtm therefore predicts a scalar confidence for each speech token using layer normalization (LN) and a sigmoid linear unit (SiLU):
\begin{equation}
  g_t = \sigma\!\left(W_2\,\operatorname{SiLU}\!\left(W_1\operatorname{LN}(a_t)\right)\right),
  \qquad g_t\in[0,1].
  \label{eq:gate}
\end{equation}
The effective weight is $w_t=m_tg_t$, where $m_t\in\{0,1\}$ is the padding mask.  We interpret $g_t$ as a learned measure of token utility for the segmentation objective, not as a calibrated probability or an estimate of acoustic corruption.  The weights modulate both the cross-attention statistics and a global audio FiLM branch.  The latter uses
\begin{equation}
  \bar a = \frac{\sum_t w_t a_t}{\sum_t w_t+\epsilon}, \qquad
  V' = V\odot(1+\gamma(\bar a))+\beta(\bar a),
\end{equation}
with zero-initialized affine projections $\gamma$ and $\beta$.  Segmentation supervision learns the confidence gate jointly with the fusion modules. Neither token-level labels nor speech transcripts are required at inference.

\subsection{Speech-Visual Kernel Linear Attention}

For one head, flatten a visual feature map into $N$ positions $v_i$ and use the exponential linear unit (ELU) to define
\begin{equation}
\begin{aligned}
  q_i&=\phi(W_qv_i),&
  k_t&=\phi(W_ka_t),&
  u_t&=W_va_t,\\
  \phi(x)&=\operatorname{ELU}(x)+1.
\end{aligned}
\end{equation}
The positive feature map permits associative normalization.  \svkla first accumulates a fixed-size speech memory
\begin{equation}
  S=\sum_{t=1}^{T} w_t k_tu_t^{\top}\in\mathbb{R}^{d\times d},
  \qquad z=\sum_{t=1}^{T} w_tk_t\in\mathbb{R}^{d},
\end{equation}
and then conditions each visual query by
\begin{equation}
  y_i=\frac{q_i^{\top}S}{q_i^{\top}z+\epsilon}.
  \label{eq:svkla}
\end{equation}
The operation computes $\phi(Q)(\phi(K)^\top U)$ and never constructs an $N\times T$ attention matrix.  Per head, conventional softmax cross-attention needs $O(NTd)$ operations and $O(NT)$ affinity memory, whereas \svkla needs $O((N+T)d^2)$ operations and $O((N+T)d+d^2)$ working storage including projected features.  The $d\times d$ sufficient statistic is especially favorable when $N$ grows with high-resolution feature maps.

\subsection{Dual-scale bilateral routing}

The $1/16$ \svkla block identifies semantically compatible regions at modest spatial cost.  Its output is upsampled and added to the $1/8$ context feature, where the second \svkla block resolves finer spatial relations.  Let $C_{16}$ and $C_{8}$ be context features, $D_8$ the detail feature, and $\mathcal{A}$ an \svkla block:
\begin{align}
  \widetilde C_{16} &= \mathcal{A}_{16}(C_{16},A,m),\\
  \widetilde C_{8} &= \mathcal{A}_{8}\!\left(C_8+\operatorname{Up}(\widetilde C_{16}),A,m\right),\\
  F_8^{0} &= \operatorname{Fuse}\!\left([D_8\,\|\,\widetilde C_8]\right), \\
  F_8 &= \operatorname{VKA}\!\left(F_8^{0}\right).
\end{align}
After a residual visual kernel-attention block, the fused $1/8$ feature is upsampled and projected to $1/4$ resolution.  Let $C_4$ denote the top-down backbone context at $1/4$, and let $D_4$ be the output of an independent shallow detail path that uses two stride-2 stages.  The high-resolution head computes
\begin{align}
  S_4 &= P_4\!\left(\operatorname{Up}(F_8)\right),\\
  R_4 &= \operatorname{Fuse}_4\!\left([C_4\,\|\,S_4\,\|\,D_4]\right),\\
  \widehat M &= \operatorname{Up}_{4\times}\!\left(H_4(R_4)\right).
\end{align}
Here, $P_4$ is a $1\times1$ semantic projection, while $\operatorname{Fuse}_4$ and $H_4$ use depthwise-separable convolutions.  This branch restores spatial detail after speech-conditioned localization without applying cross-modal attention at the much larger $1/4$ grid.  We optimize binary cross-entropy (BCE) plus soft Dice loss:
\begin{equation}
  \mathcal{L}=\mathcal{L}_{\mathrm{BCE}}+
  1-\frac{2\sum_p \sigma(\widehat m_p)m_p+1}
  {\sum_p\sigma(\widehat m_p)+\sum_pm_p+1}.
\end{equation}

\section{Experiments}
\label{sec:experiments}

\subsection{Experimental protocol}

We resize images and masks to 352$\times$352, cap audio at 8 s, and use frozen wav2vec~2.0 base with 256-dimensional projected tokens.  Visual backbones use ImageNet initialization.  Both reported \model variants enable the $1/4$ high-resolution refinement head.  All models train for 40 epochs with AdamW, batch size 8, weight decay $10^{-4}$, cosine annealing, automatic mixed precision, and gradient-norm clipping at 1.0.  Newly initialized audio-fusion and decoder parameters use a learning rate of $2\times10^{-4}$.  For every Swin-Base model, the pretrained backbone is optimized separately at $2\times10^{-5}$. Training uses the one-voice-per-source-per-epoch sampler in \cref{eq:voice_sampling}.

\begin{table*}[t]
\centering
\caption{Overall accuracy on the hard test split (\%).  Each clean-selected checkpoint is evaluated once per source under its assigned noise-type--severity condition.  The best and second-best values are shown in bold and underlined, respectively.}
\label{tab:hard_overall}
\footnotesize
\setlength{\tabcolsep}{3.1pt}
\begin{tabular}{llrrrrrrr}
\toprule
Method & Backbone & mIoU & oIoU & Pr@.5 & Pr@.6 & Pr@.7 & Pr@.8 & Pr@.9 \\
\midrule
Audio U-Net & CNN & 22.33 & 29.06 & 16.63 & 11.70 & 7.99 & 4.72 & 1.51 \\
Audio FCN & ResNet-101 & 37.79 & 48.89 & 41.30 & 35.62 & 29.10 & 20.81 & 9.40 \\
Audio DeepLabV3 & ResNet-101 & 42.48 & 53.32 & 46.51 & 41.13 & 34.59 & 25.36 & 12.29 \\
\midrule
FIANet adapter~\cite{lei2025fianet} & Swin-B & 46.37 & 57.46 & 50.55 & 45.23 & 39.11 & 30.68 & 17.40 \\
SBANet adapter~\cite{li2025sbanet} & Swin-B & 48.03 & \underline{58.95} & 52.81 & 47.17 & 40.58 & 31.54 & 17.56 \\
STDNet adapter~\cite{zhang2025stdnet} & Swin-B & 44.57 & 52.70 & 46.50 & 40.84 & 34.64 & 26.54 & 15.00 \\
LSCF adapter~\cite{ma2025lscf} & Swin-B & 48.46 & 58.91 & 53.20 & 47.73 & 41.11 & 32.37 & 18.10 \\
\midrule
\model-ResNet101 (ours) & ResNet-101 & \underline{50.86} & 57.68 & \underline{56.22} & \underline{51.10} & \underline{45.15} & \underline{36.00} & \underline{20.59} \\
\textbf{\model-Swin-B (ours)} & Swin-B & \textbf{54.09} & \textbf{60.81} & \textbf{59.89} & \textbf{54.84} & \textbf{48.25} & \textbf{39.17} & \textbf{23.22} \\
\bottomrule
\end{tabular}
\end{table*}

The primary model is selected by validation mIoU.  Test masks are generated at 512$\times$512 with a probability threshold of 0.5.  The shared metrics are overall intersection over union (oIoU), mean intersection over union (mIoU), and precision at threshold $\tau$ (Pr@$\tau$):
\begin{align}
  \operatorname{oIoU} &= \frac{\sum_j|\widehat M_j\cap M_j|}{\sum_j|\widehat M_j\cup M_j|},\\
  \operatorname{mIoU} &= \frac{1}{J}\sum_j\frac{|\widehat M_j\cap M_j|}{|\widehat M_j\cup M_j|},\\
  \operatorname{Pr@}\tau &= \frac{1}{J}\sum_j \mathbf{1}[\operatorname{IoU}_j\geq\tau].
\end{align}
We report Pr@$\tau$ for $\tau\in\{0.5,0.6,0.7,0.8,0.9\}$.  All methods are evaluated on the same clean and hard spoken-query test splits with no hard-set fine-tuning; no text-only RISBench result is reused.

\subsection{Baselines}

We compare against three established segmentation architectures and four recent RRSIS models.  Audio U-Net applies feature-wise linear modulation (FiLM) at its bottleneck; FCN-ResNet101 and DeepLabV3-ResNet101 apply the same pooled-audio FiLM conditioning to their high-level backbone feature~\cite{ronneberger2015unet,long2015fcn,chen2017deeplab}.  These controlled baselines share the waveform encoder and isolate the effect of the visual segmentation architecture.

FIANet, SBANet, STDNet, and LSCF were originally proposed for text-guided RRSIS with Swin-Base visual features~\cite{lei2025fianet,li2025sbanet,zhang2025stdnet,ma2025lscf,liu2021swin}.  Their entries here are explicitly \emph{audio adaptations}, not numbers copied from the authors' text-input papers: the text tokens are replaced with our shared wav2vec~2.0 tokens while retaining, respectively, fine-grained alignment and multi-scale enhancement, scale-wise bidirectional alignment, spatial multi-scale correlation with target/background decoding, and long-term semantic guidance.  Finally, \model-ResNet101 is a backbone control for the primary Swin-Base model; both use the same detail path, dual-scale \svkla, \cgtm, and decoder.

\paragraph{Quantitative comparison}
\Cref{tab:main_accuracy} shows that \model-Swin-B ranks first in every reported metric.  Against LSCF, the strongest non-ours method, it improves mIoU from 56.71\% to 62.09\% and oIoU from 66.14\% to 68.22\%, absolute gains of 5.38 and 2.08 percentage points (9.49\% and 3.15\% relative).  The gains at Pr@0.5, 0.6, and 0.7 are 5.92, 6.70, and 7.03 points, showing that the improvement is not driven only by a small number of high-IoU examples.  Among methods sharing Swin-Base, the mIoU margins over LSCF, SBANet, FIANet, and STDNet are 5.38, 6.81, 8.61, and 11.75 points, respectively.  Because the visual backbone and speech encoder are controlled in this comparison, these gains are consistent with the benefit of token-preserving dual-scale fusion, global kernel attention, and high-resolution refinement rather than a stronger encoder alone.

The backbone control provides complementary evidence.  Replacing ResNet-101 with Swin-Base while keeping the AeroReformer2 fusion and decoder fixed raises mIoU by 2.83 points and oIoU by 1.99 points.  Hierarchical shifted-window features appear to complement the top-down context path in scenes with repeated objects and long-range spatial phrases.  The large gap to U-Net (34.25 mIoU points) and the 10.22-point gain over DeepLabV3 further indicate that bottleneck-level pooled conditioning is insufficient for instance-specific spoken grounding.

\paragraph{High-overlap behavior}
The $1/4$ refinement head particularly improves the high-overlap regime.  \model reaches 46.15\% at Pr@0.8 and 27.49\% at Pr@0.9, exceeding LSCF by 6.48 and 5.30 points, respectively.  The refinement stage is applied only after token-conditioned localization, so these boundary gains do not require an $N\times T$ speech--visual affinity map at $1/4$ resolution.

\subsection{Hard-set robustness}
\label{sec:hard_results}

\begin{table}[t]
\centering
\caption{Hard-test mean intersection over union (mIoU) by interference type and severity for the Swin-Base methods (\%).  Low, medium, and high correspond to nominal signal-to-noise ratios (SNRs) of 15, 7.5, and 0 decibels (dB).}
\label{tab:hard_grid}
\footnotesize
\setlength{\tabcolsep}{2.6pt}
\begin{tabular}{lrrrrrrrrr}
\toprule
& \multicolumn{3}{c}{Rotor} & \multicolumn{3}{c}{Wind} & \multicolumn{3}{c}{Mixed} \\
\cmidrule(lr){2-4}\cmidrule(lr){5-7}\cmidrule(lr){8-10}
Method & Low & Med. & High & Low & Med. & High & Low & Med. & High \\
\midrule
FIANet adapter & 52.31 & 49.27 & 37.70 & 52.42 & 47.96 & 38.55 & 54.16 & 47.75 & 37.18 \\
SBANet adapter & 54.40 & 51.78 & 37.73 & 53.05 & 51.13 & 39.74 & 55.20 & 51.28 & 37.92 \\
STDNet adapter & 49.96 & 47.82 & 36.06 & 49.64 & 47.59 & 37.99 & 49.98 & 46.27 & 35.83 \\
LSCF adapter & 55.57 & 52.49 & 36.73 & 55.71 & 51.23 & 38.94 & 56.39 & 51.58 & 37.52 \\
\textbf{\model} & \textbf{61.94} & \textbf{58.36} & \textbf{41.43} & \textbf{61.55} & \textbf{57.42} & \textbf{44.76} & \textbf{61.27} & \textbf{57.36} & \textbf{42.70} \\
\bottomrule
\end{tabular}
\end{table}

\Cref{tab:hard_overall} shows that \model-Swin-B remains the strongest method under the complete hard-test mixture and now ranks first in every metric.  It exceeds LSCF, the highest-mIoU non-ours model, by 5.63 mIoU points and improves oIoU by 1.86 points over SBANet, the strongest non-ours method on that metric.  The margins over LSCF are 6.69 points at Pr@0.5 and 5.12 points at Pr@0.9, indicating that the advantage extends from reliable referent selection to high-overlap masks.

Relative to clean evaluation, \model-Swin-B loses 8.00 mIoU and 7.41 oIoU points.  Thus, despite the improved absolute hard-set accuracy, acoustic interference remains a material limitation.

The condition breakdown in \cref{tab:hard_grid} provides a more detailed view.  \model-Swin-B ranks first in mIoU in all nine cells, with margins over the best baseline ranging from 3.70 to 6.37 points.  The gains remain 3.70, 5.02, and 4.78 points at high-severity rotor, wind, and mixed interference, respectively.  All methods nevertheless deteriorate sharply near 0 dB, indicating that the benchmark does not reduce to a benign augmentation test.  Because no hard-set ablation of the confidence gate is available, these results characterize the complete architecture and should not be interpreted as evidence that \cgtm estimates or suppresses interference.

\subsection{Ablation study}

As shown in \cref{tab:ablation}, post-fusion kernel attention is the most influential tested component: removing it reduces mIoU by 4.50 points and oIoU by 1.37 points, with losses of 5.69, 5.14, and 2.60 points at Pr@0.5, Pr@0.7, and Pr@0.9, respectively.  This block operates after the detail and speech-conditioned context streams have been merged, so its global refinement can suppress spatially disconnected distractors while restoring coherence to the selected object.  Removing the $1/32$ visual KLA causes smaller but consistent drops of 1.01 mIoU and 0.24 oIoU, indicating that global low-resolution context complements the two cross-modal fusion stages.  Replacing the learned confidence gate with uniform token weights lowers mIoU by 0.27 points and oIoU by 0.01 points; its larger 1.29-point reduction at Pr@0.9 indicates that confidence weighting contributes most clearly to stringent-overlap predictions.  A matched ablation of the $1/4$ head is not available in the current experiment set.

\subsection{Qualitative comparisons}

\begin{figure*}[t]
\centering
\includegraphics[width=\linewidth]{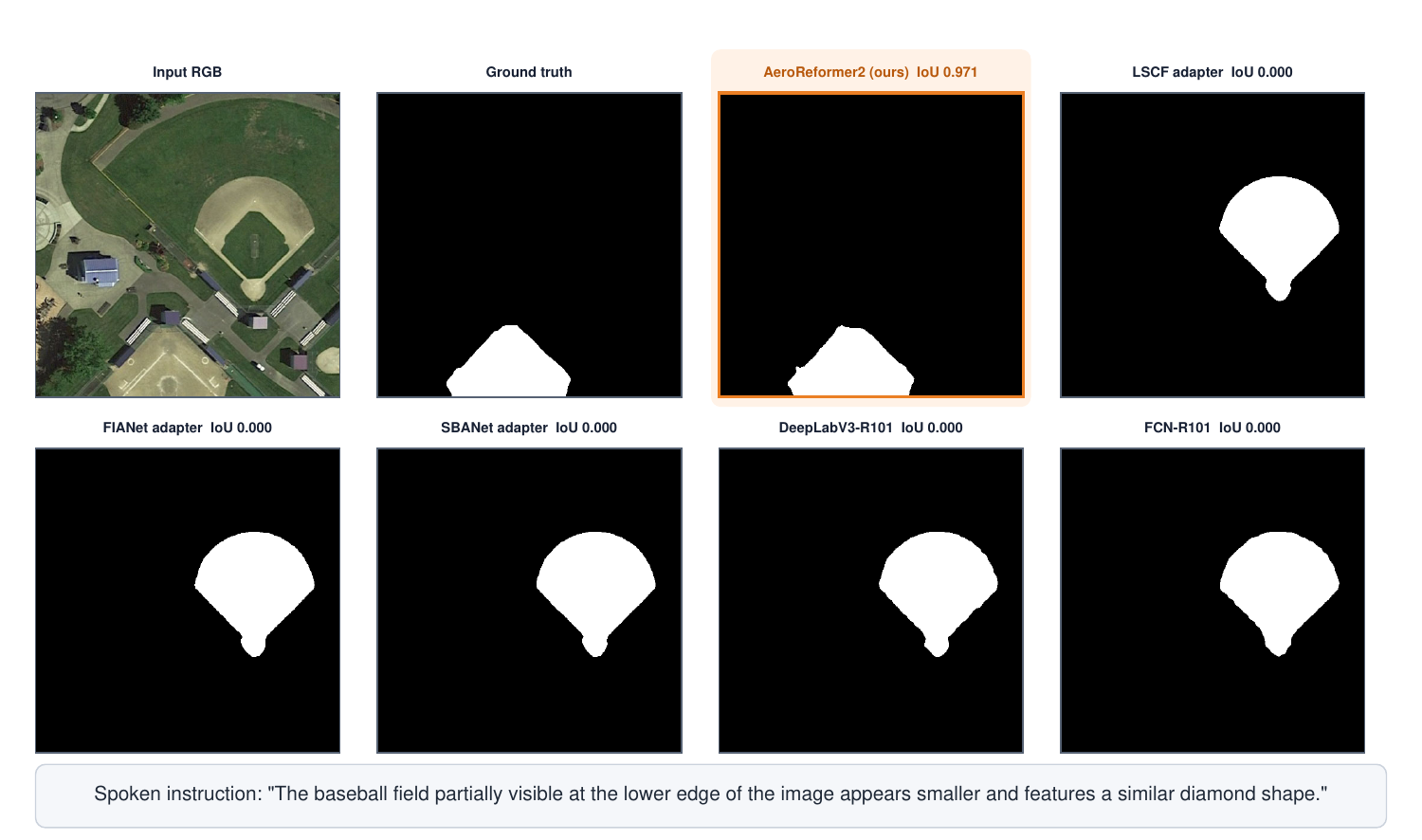}
\caption{Baseball-field example.  The instruction selects the smaller field that is only partially visible at the lower image boundary.  \model recovers that truncated diamond with 0.971 IoU, whereas all five comparison methods select the fully visible central field and therefore obtain zero overlap.}
\label{fig:qualitative_baseball}
\end{figure*}

\begin{figure*}[t]
\centering
\includegraphics[width=\linewidth]{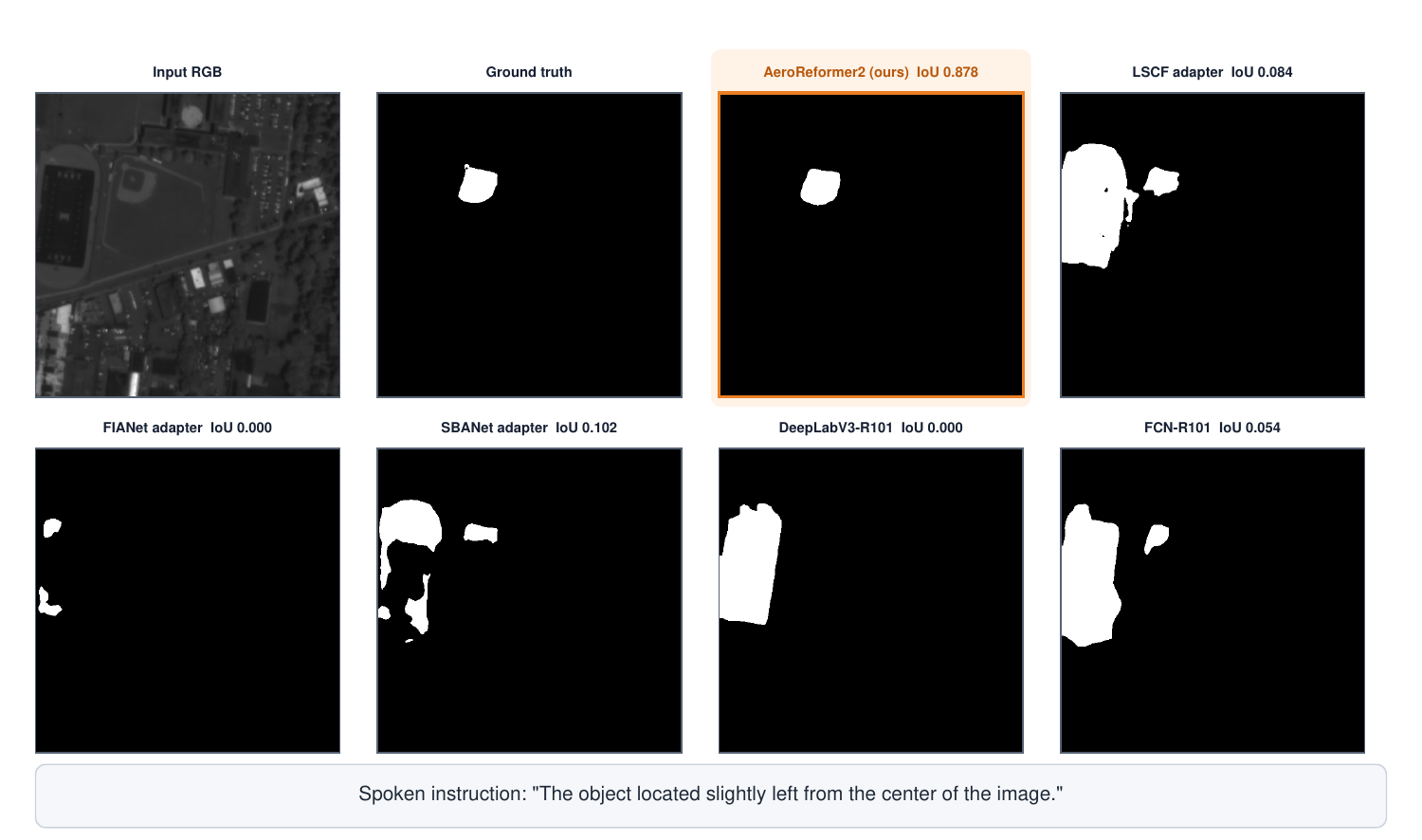}
\caption{Small-object example.  The instruction refers to the object slightly left of image center.  \model isolates the compact target with 0.878 IoU, whereas the alternatives activate on larger structures, image boundaries, or multiple distractors.}
\label{fig:qualitative_object}
\end{figure*}

\begin{figure*}[t]
\centering
\includegraphics[width=\linewidth]{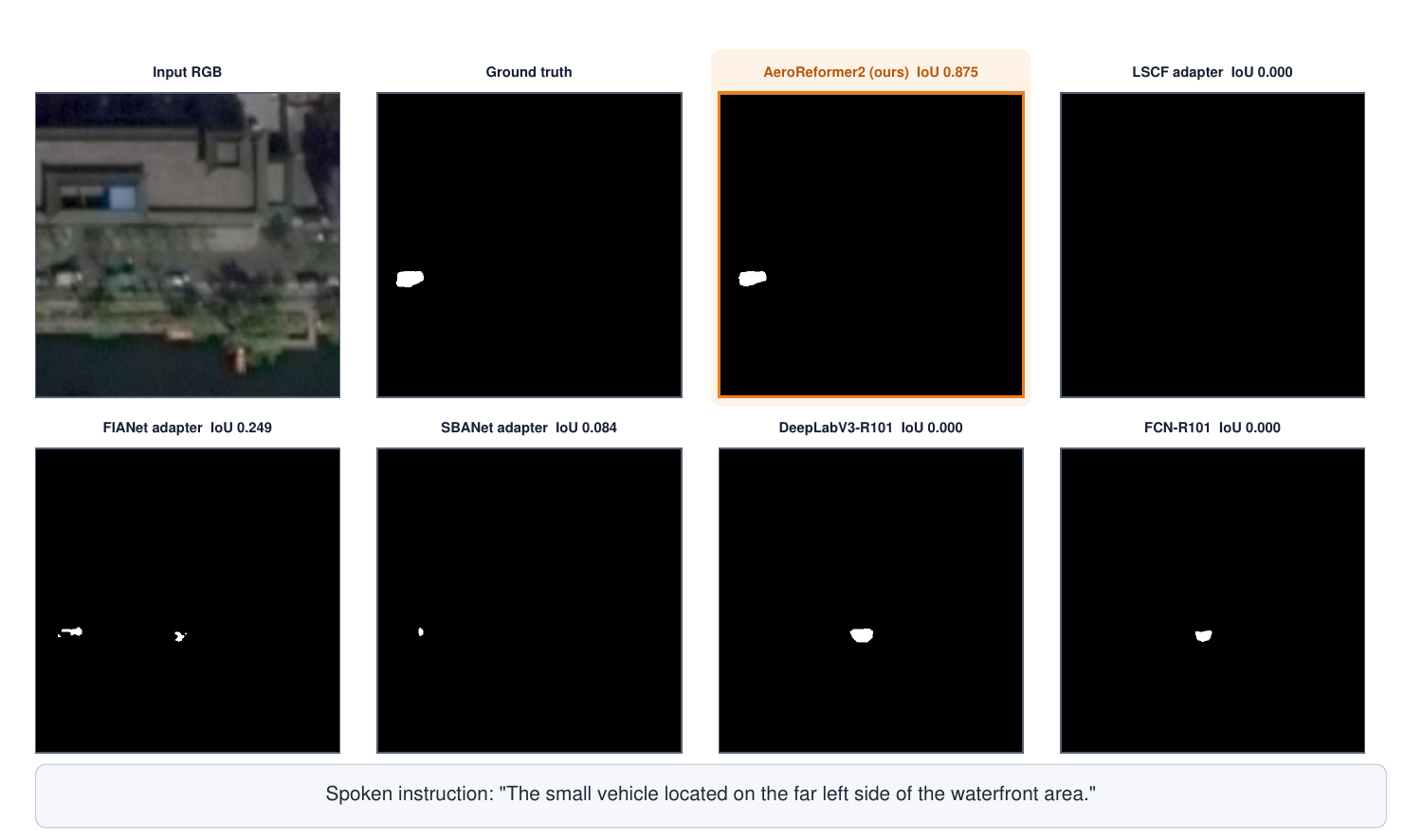}
\caption{Waterfront-vehicle example.  The target is a very small vehicle on the far-left side of the waterfront.  \model preserves the correct location and compact extent with 0.875 IoU; the comparison methods miss it, fragment it, or select a different object.}
\label{fig:qualitative_vehicle}
\end{figure*}

\Cref{fig:qualitative_baseball,fig:qualitative_object,fig:qualitative_vehicle} compare six methods on examples involving a partially visible sports field, a small central object, and a tiny waterfront vehicle.  Each panel contains the aerial image, ground-truth mask, and predicted masks at the native 512$\times$512 evaluation resolution.

\begin{table}[t]
\centering
\caption{Component ablation of \model-Swin-B on the clean test split (\%). Each ablation row removes one component from the complete model.}
\label{tab:ablation}
\footnotesize
\setlength{\tabcolsep}{3.8pt}
\begin{tabular}{lrrrrr}
\toprule
Variant & mIoU & oIoU & Pr@.5 & Pr@.7 & Pr@.9 \\
\midrule
w/o post-fusion KLA & 57.59 & 66.85 & 63.56 & 51.57 & 24.89 \\
w/o $1/32$ visual KLA & 61.08 & 67.98 & 68.40 & 56.03 & 26.84 \\
w/o confidence gate & 61.82 & 68.21 & 69.16 & 56.52 & 26.20 \\
\midrule
\textbf{Full \model} & \textbf{62.09} & \textbf{68.22} & \textbf{69.25} & \textbf{56.71} & \textbf{27.49} \\
\bottomrule
\end{tabular}
\end{table}

\paragraph{Partially visible baseball field}
In \cref{fig:qualitative_baseball}, \model reaches 0.971 IoU and follows the visible portion of the lower-edge field.  Every comparison method instead segments the salient central field and obtains zero IoU.  The example isolates relational grounding: recognizing the object category and shape is insufficient unless ``partially visible at the lower edge'' is used to select the intended instance.

\begin{table*}[t]
\centering
\caption{End-to-end efficiency on an RTX 4070 Laptop graphics processing unit (GPU) at 352$\times$352, batch size 1, 8-second audio, and automatic mixed precision (AMP). Latency is the median of three synchronized 50-iteration blocks after 10 warm-up iterations. Parameters and memory include the frozen wav2vec~2.0 encoder. M denotes millions, FPS denotes frames per second, and MB denotes megabytes.}
\label{tab:efficiency}
\small
\setlength{\tabcolsep}{3.8pt}
\begin{tabular}{lrrrrrr}
\toprule
Method & Params (M) & Trainable (M) & Latency (ms) & FPS & Peak (MB) & mIoU \\
\midrule
Audio U-Net & 96.627 & 2.255 & 15.41 & 64.91 & 546.6 & 27.84 \\
Audio FCN-R101 & 147.560 & 53.188 & 24.92 & 40.12 & 758.8 & 49.54 \\
Audio DeepLabV3-R101 & 154.247 & 59.875 & 27.51 & 36.35 & 804.1 & 51.87 \\
\midrule
FIANet adapter & 182.428 & 88.056 & 38.04 & 26.29 & 928.9 & 53.48 \\
SBANet adapter & 182.904 & 88.533 & 39.99 & 25.00 & 948.7 & 55.28 \\
STDNet adapter & 183.169 & 88.797 & 42.22 & 23.69 & 969.7 & 50.34 \\
LSCF adapter & 184.936 & 90.565 & 44.00 & 22.73 & 991.5 & 56.71 \\
\midrule
\model-R101 (ours) & 138.007 & 43.635 & 28.67 & 34.88 & 666.9 & 59.26 \\
\textbf{\model-Swin-B (ours)} & \textbf{182.063} & \textbf{87.692} & 39.19 & 25.52 & \textbf{855.6} & \textbf{62.09} \\
\bottomrule
\end{tabular}
\end{table*}

\paragraph{Small object near the center}
In \cref{fig:qualitative_object}, \model obtains 0.878 IoU and closely matches the compact target.  The strongest alternative reaches only 0.102 IoU; the remaining predictions range from 0 to 0.084 and mostly cover large peripheral structures or disconnected regions.  This case shows the benefit of retaining the directional phrase ``slightly left from the center'' when appearance alone provides a weak target cue.

\paragraph{Tiny waterfront vehicle}
In \cref{fig:qualitative_vehicle}, \model achieves 0.875 IoU on a target occupying only a small fraction of the image.  FIANet reaches 0.249 but adds a second false-positive fragment, while SBANet reaches 0.084 and the other three methods obtain zero overlap.  The comparison illustrates the complementary roles of far-left directional grounding and the $1/4$ detail refinement head for tiny-object delineation.

\subsection{Efficiency analysis}

\paragraph{Hardware-matched end-to-end profile}
We profile all nine models in \cref{tab:main_accuracy} on an NVIDIA GeForce RTX 4070 Laptop graphics processing unit (GPU) with 8 gigabytes (GB) of memory, using PyTorch 2.13 and NVIDIA Compute Unified Device Architecture (CUDA) 13.0.  Every model receives a 352$\times$352 red--green--blue (RGB) tensor and an 8-second, 16-kHz waveform at batch size one under automatic mixed precision (AMP).  The frozen wav2vec~2.0 encoder is included in the parameter count, latency, throughput, and peak allocated memory.  Latency is measured after 10 warm-up iterations using three blocks of 50 synchronized iterations, and the median block latency is reported.

\paragraph{Accuracy--efficiency comparison}
As shown in \cref{tab:efficiency}, the primary Swin-Base model gives the highest accuracy rather than the lowest absolute latency. Among the five Swin-Base systems, it has the fewest total and trainable parameters and the lowest peak memory. Relative to LSCF, it is 10.9\% faster (39.19 versus 44.00 ms), uses 13.7\% less peak memory, and improves mIoU by 5.38 points. Relative to FIANet, it has 3.0\% higher latency but gains 8.61 mIoU points while reducing peak memory by 7.9\%. Relative to SBANet, it is 2.0\% faster, improves mIoU by 6.81 points, and uses 9.8\% less peak memory. It is also 7.2\% faster than STDNet, uses 11.8\% less memory, and is 11.75 mIoU points more accurate. The efficiency advantage is therefore expressed in both memory use and accuracy per model capacity, although the complete Swin-Base network is not the fastest model in every comparison.

The ResNet-101 version provides a second operating point: it retains 59.26\% mIoU at 28.67 ms (34.88 FPS) and 666.9 MB. Compared with \model-Swin-B, it sacrifices 2.83 mIoU points but lowers latency and peak memory by 26.8\% and 22.1\%, respectively. The conventional baselines are faster, but their pooled audio conditioning incurs substantial accuracy losses: DeepLabV3 is 7.39 mIoU points lower, and U-Net is 31.42 points lower despite its higher throughput.

\paragraph{Analytical attention state}
At 352$\times$352, the $1/16$ and $1/8$ maps contain 484 and 1,936 positions.  An 8-second wav2vec-like sequence at 50 tokens/s has $T=400$.  With four heads and $d=16$, softmax attention across both scales materializes 3,872,000 affinity values (14.771 mebibytes (MiB) in 32-bit floating-point (FP32) format).  \svkla stores 2,048 head-specific $d\times d$ state values (0.007812 MiB), a 1,890.62$\times$ reduction in the cross-modal affinity state.  The corresponding simplified multiply--accumulate (MAC) counts are 123.904 million (M) and 3.297 M, respectively, or 37.58$\times$ lower for the attention core.  Common query, key, and value projections and activation tensors are excluded from both sides.  At 512$\times$512, the same calculation yields a 4,000$\times$ affinity-state and 43.24$\times$ attention-core arithmetic reduction (\cref{tab:complexity}).

\paragraph{Scope of the efficiency claim}
The measured profile and the analytical attention analysis answer different questions.  \Cref{tab:efficiency} includes the complete audio encoder, visual backbone, fusion modules, and decoder, whereas \cref{tab:complexity} isolates the cross-modal attention core.  Optimized convolution and softmax kernels can make a model with denser attention faster for a particular tensor shape, so the linear formulation guarantees neither the lowest end-to-end latency nor the same ranking on another device.  It does, however, remove the $N\times T$ affinity state, which is consistent with the lower peak memory of \model among the Swin-Base comparisons.

\begin{table*}[t]
\centering
\caption{Analytical cross-modal attention-core complexity for an 8-second, 400-token utterance, four heads, and $d=16$. MiB denotes mebibytes, M denotes millions, and MACs denotes multiply--accumulate operations. These are not end-to-end memory or latency measurements.}
\label{tab:complexity}
\small
\begin{tabular}{rrrrrr}
\toprule
Image & Softmax state & Linear state & State reduction & Softmax MACs & MAC reduction \\
\midrule
352 & 14.771 MiB & 0.007812 MiB & 1,890.62$\times$ & 123.904 M & 37.58$\times$ \\
512 & 31.250 MiB & 0.007812 MiB & 4,000.00$\times$ & 262.144 M & 43.24$\times$ \\
\bottomrule
\end{tabular}
\end{table*}

\section{Conclusion and Future Work}
\label{sec:conclusion}

This paper addresses the missing interaction layer between spoken user intent and pixel-level remote-sensing perception.  \dataset converts a large text-based RRSIS resource into a controlled audio benchmark with eight training voices, balanced single-voice clean evaluation, and a nine-condition hard evaluation grid.  Its epoch sampler exposes one acoustic realization per source image--mask pair per epoch, preserving the statistical size of the visual training set.  The linguistic analysis shows that the task is heavily driven by position, size, and relational cues, which supports retaining speech tokens rather than relying only on a pooled utterance vector.

\model is designed for this setting.  Its bilateral paths balance semantic context with small-object boundaries; \svkla performs token-level grounding without an explicit visual--speech affinity matrix; \cgtm assigns task-dependent confidence before token aggregation; dual-scale fusion routes speech to coarse localization and fine grounding; and the $1/4$ refinement head restores high-resolution boundaries after fusion.  On the clean test split, \model-Swin-B reaches 62.09\% mIoU and 68.22\% oIoU, improving over the strongest non-ours method by 5.38 and 2.08 percentage points.  It also attains the highest hard-set mIoU of 54.09\%, a 5.63-point margin over LSCF.  In the component ablation, post-fusion kernel attention contributes 4.50 mIoU points over its removal.  On the matched RTX 4070 profile, the full model uses the least memory among Swin-Base methods, is 10.9\% faster and 13.7\% lighter in peak memory than LSCF, and offers a 28.67-ms ResNet-101 operating point.

Future work should collect synchronized speech from remote-sensing analysts and source-matched aerial or satellite imagery; evaluate realistic office, control-room, outdoor, radio, and microphone conditions; extend the benchmark to multilingual and code-switched commands; model streaming speech for low-latency online masks; calibrate uncertainty and allow clarification when a command is ambiguous; and compress the audio encoder for efficient deployment.  Cross-dataset evaluation across satellite and airborne image sources, together with human-in-the-loop studies, will be essential to establish whether improved benchmark metrics translate into faster and more reliable image-analysis workflows.

\section*{Declaration of Competing Interest}
\par The authors declare that they have no known competing financial interests or personal relationships that could have appeared to influence the work reported in this paper.

\bibliographystyle{elsarticle-num}
\bibliography{references}

\end{document}